\documentclass[letterpaper, 10 pt, journal, twoside]{IEEEtran}

\IEEEoverridecommandlockouts                              

\usepackage{amssymb}

\usepackage[bookmarks=true]{hyperref}
\usepackage{amsmath}
\usepackage{color,soul}
\usepackage{xcolor}
\usepackage{graphicx}

\usepackage{threeparttable}

\title{
Grasp Transfer based on  Self-Aligning \\ Implicit Representations of Local Surfaces}

\author{Ahmet Tekden$^{1}$, Marc Peter Deisenroth$^{2}$, and Yasemin Bekiroglu$^{1,2}$
\thanks{Manuscript received: March 31, 2023; Revised June 28, 2023; Accepted July 27, 2023.}
\thanks{This paper was recommended for publication by Editor Hong Liu upon evaluation of the Associate Editor and Reviewers' comments.}
\thanks{This work was supported by Chalmers AI Research Center (CHAIR) and Chalmers Gender Initiative for Excellence (Genie), the project AIMCoR - AI-enhanced Mobile Manipulation Robot for Core Industrial Applications and partially supported by the Wallenberg AI, Autonomous Systems and Software Program (WASP) funded by the Knut and Alice Wallenberg Foundation.}
\thanks{$^{1}$Electrical Engineering Department, Chalmers University of Technology, Sweden
        (email:{\tt\footnotesize tekden@chalmers.se}).}%
\thanks{$^{2}$Computer Science Department, University College London, U.K.}%

\thanks{Digital Object Identifier (DOI): see top of this page.}
}

\begin{document}

\maketitle

\begin{abstract}
Objects we interact with and manipulate often share similar parts, such as handles, that allow us to transfer our actions flexibly due to their shared functionality. This work addresses the problem of transferring a grasp experience or a demonstration to a novel object that shares shape similarities with objects the robot has previously encountered. Existing approaches for solving this problem are typically restricted to a specific object category or a parametric shape. Our approach, however, can transfer grasps associated with implicit models of local surfaces shared across object categories. Specifically, we employ a single expert grasp demonstration to learn an implicit local surface representation model from a small dataset of object meshes. At inference time, this model is used to transfer grasps to novel objects by identifying the most geometrically similar surfaces to the one on which the expert grasp is demonstrated. Our model is trained entirely in simulation and is evaluated on simulated and real-world objects that are not seen during training. Evaluations indicate that grasp transfer to unseen object categories using this approach can be successfully performed both in simulation and real-world experiments. The simulation results also show that the proposed approach leads to better spatial precision and grasp accuracy compared to a baseline approach. 

\end{abstract}
\section{Introduction}
\IEEEPARstart{G}{rasp} synthesis has been studied extensively~\cite{grasp_synthesis_review} as robotic grasping skills have a significant impact on the success of subsequent manipulation tasks. A common approach for generating grasps is to train a prediction network, which either takes top-down scene images~\cite{handeye_grasp, transporter_net} or 3D point clouds~\cite{acronym, ten2017grasp, contactgraspnet, 6dofgrasp, giga} for predicting multiple grasp locations. However, such models are data-hungry, and have to use non-curated data since data annotation at scale is often not feasible in robotics. Another approach for generating grasp poses is based on finding category-level correspondences~\cite{kpam,NeRF-Supervision,NDF} and using them for grasp prediction. Compared to grasp prediction networks, correspondence-based methods are more data-efficient and allow for transferring grasp demonstrations to other objects, thereby facilitating learning from demonstration~\cite{lfd_survey}, and imitation learning~\cite{imitation_learning_review}. However, correspondence-based methods can predict correct grasp locations only on the trained object categories. 

\begin{figure}[t]
	\centering
    \includegraphics[width=0.80\linewidth]{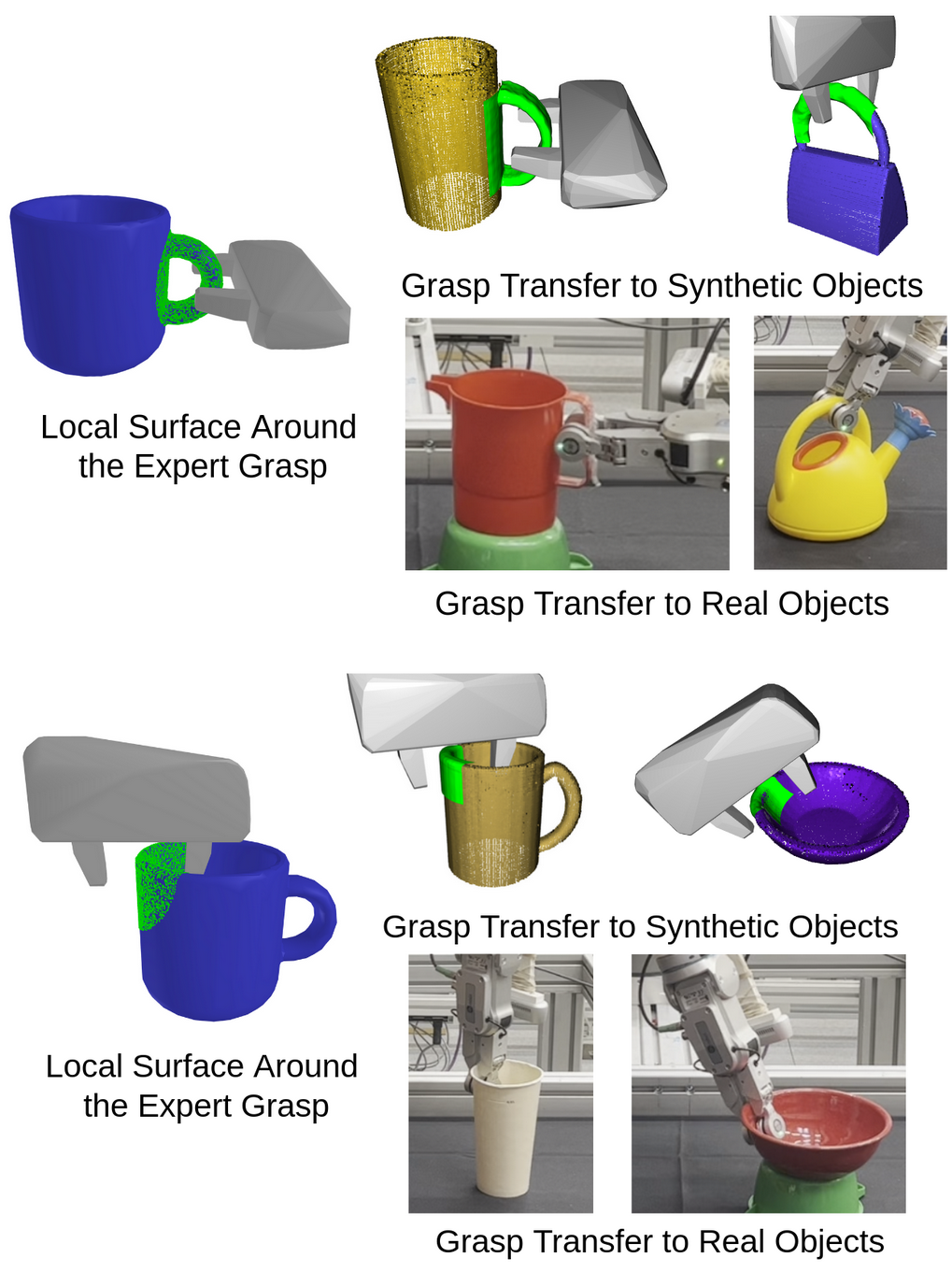}
    \caption{
    Example experiments of grasp transfer to various objects. Utilizing implicit representations of local surfaces, grasps are transferred to novel objects by identifying local surfaces that are geometrically similar to those surrounding the expert grasps. 
    }
    \label{fig:page1}
\end{figure} 

Compared to approaches for synthesizing grasp candidates from scratch~\cite{refinement, qin2020s4g} or relying on object-level correspondences~\cite{kpam,NeRF-Supervision,NDF}, making use of previous experiences/demonstrations from similar objects and transferring grasps onto new object instances can be more data efficient~\cite{prototypepart} and facilitate task-oriented grasping~\cite{taskOriented}. In this study, we address the problem of reusing grasping knowledge when faced with a similar object without planning from scratch. Specifically, we focus on transferring grasps across object categories that share geometrically similar surfaces/parts, as illustrated in Figure~\ref{fig:page1}. Using our method, we can match the local surfaces learned from, e.g. the handle or the rim of a mug object, to identify the corresponding grasp location on other objects such as bags and bowls that share similar parts. 

In this work, we model local surfaces with implicit representations using neural fields~\cite{nerf_survey} and perform shape inference by optimizing latent shape codes. Neural fields are position-based neural networks that take continuous spatial coordinates as input to parametrize a variety of functions, e.g. implicit surfaces~\cite{deepsdf,occ-net}. For latent shape code optimization, most neural field methods require all objects to be in canonical reference frames both at training and inference time~\cite{deepsdf}. Aligning objects to the exact canonical poses is challenging as it requires the prediction of object poses~\cite{nocs, 6pack}, which often introduces a certain degree of error. We propose a novel approach for aligning object or surface poses to a common reference frame through pose embedding optimization. We rely on a small dataset of object models from a single category and a single grasp demonstration on one object within that category. By using the demonstrated grasp and the corresponding surface around it as a reference, we extract geometrically similar surfaces from the object models. These are then used to train an implicit surface representation. At inference time, the learned implicit representations are employed to identify surfaces on novel objects, based on the ones that they are trained on. This is achieved without exact pose knowledge and enables us to transfer grasps across different objects. The main contributions of this work can be summarized as follows:
\begin{itemize}
 \item We present an end-to-end surface reconstruction approach that simultaneously learns to reconstruct, scale, and align shapes in a self-supervised manner. The proposed method facilitates learning implicit representations without complete surface match and alignment. 
 \item We introduce a novel approach for learning local surface models using a sampling sphere placed on a provided reference frame. This sphere is used for sampling the training data and by adapting it to the current alignment pose, the local surface is modeled. This in turn allows for learning the local implicit surfaces on explicitly provided locations via demonstrated grasps.
 \item Our approach successfully transfers grasp demonstrations from a single object to other objects, including ones from unseen categories. We utilize learned implicit representations to identify and localize local surfaces on test objects that are geometrically similar to those surrounding the demonstrated grasps, thereby enabling the transfer of grasp poses.
 \item In our results, we show that our method acquires better spatial precision and grasp accuracy compared to a state of the art baseline method. In addition, it acquires competitive performance on grasp transfer to novel objects in both simulation and real-world experiments.
 \end{itemize}
 
\section{Related Work}

Early works on grasp prediction that rely on explicit shape information are usually built on primitive-based correspondences~\cite{primitive_shape_grasp_1,primitive_shape_grasp_2}, parametrization using smooth differentiable functions (e.g. Grasp Moduli Spaces~\cite{pokorny2013grasp,pokorny2014grasp}), or hard-coded geometric features~\cite{myers2015affordance}. Primitive-based methods, in general, can struggle to generalize to objects that deviate significantly from the provided primitives and parametrization can deteriorate with partial point cloud data. Additionally, most other geometric feature-based approaches require explicit design and engineering effort. In contrast, our method learns local surface representations from surfaces extracted from an object dataset, enabling it to better generalize to shape deviations. Furthermore, by optimizing a latent shape code through shape reconstruction, our method can utilize partial point clouds of objects.

Various representations have been proposed for building models related to object parts to be used for grasp planning~\cite{prototypepart,kroemer2012kernel}. \cite{prototypepart} uses prototype parts to transfer grasps across novel object categories using a similarity measure to identify parts with similar shapes. However, these methods do not explicitly model local object parts, which can limit their ability to capture important shape details. In contrast, our approach explicitly models local surfaces, allowing it to capture more information about the underlying shapes.

Another approach for grasp transfer is to use category-level key points to transfer grasps to objects in the same category~\cite{kpam,NeRF-Supervision,NDF,florence2018dense,nodeslam,catgrasp}. However, such keypoint-based methods are category driven, and by design, they cannot be utilized to transfer demonstrated grasps to novel object categories. Instead of modeling object-level correspondences, our network learns implicit representations of local surfaces. For representing each different surface type, we train individual shape reconstruction models. Our method leverages these models to identify grasp poses corresponding to each model's local surface type. When the encountered surface resembles the local surfaces the model is trained on, it can accurately reconstruct its shape. However, when presented with a surface that deviates significantly from what the model is trained on, the model will fail to reconstruct its shape, leading to higher errors. Employing these local surface models enables our method to successfully transfer grasps to novel objects. 

We model local surfaces by leveraging coordinate-based neural networks, which have proven effective in representing 3D shapes and scenes in prior research~\cite{deepsdf,occ-net,siren,dif-net,NeRF}. However, these networks do not address the task of modeling local surfaces. To achieve this, we utilize a single grasp demonstration to extract geometrically similar local surfaces from a dataset of an object category. However, due to geometric variations among the objects, the poses of extracted local surfaces are not properly aligned to the reference frame of the original grasp demonstration. The proposed network architecture addresses this issue by aligning the poses of these surfaces to the desired reference frame while simultaneously learning to reconstruct their shapes. Recently it has been shown that camera poses can be estimated through neural radiance field (NeRF) training~\cite{iNeRF, barf}. However, these methods are limited to images captured from a single scene that contain patches with matching features that guide the alignment process. In contrast, our approach aligns poses of local surfaces coming from different objects by leveraging their geometric similarities.  

\section{Method}

\begin{figure*}[t]
     \centering
    \includegraphics[width=0.85\linewidth]{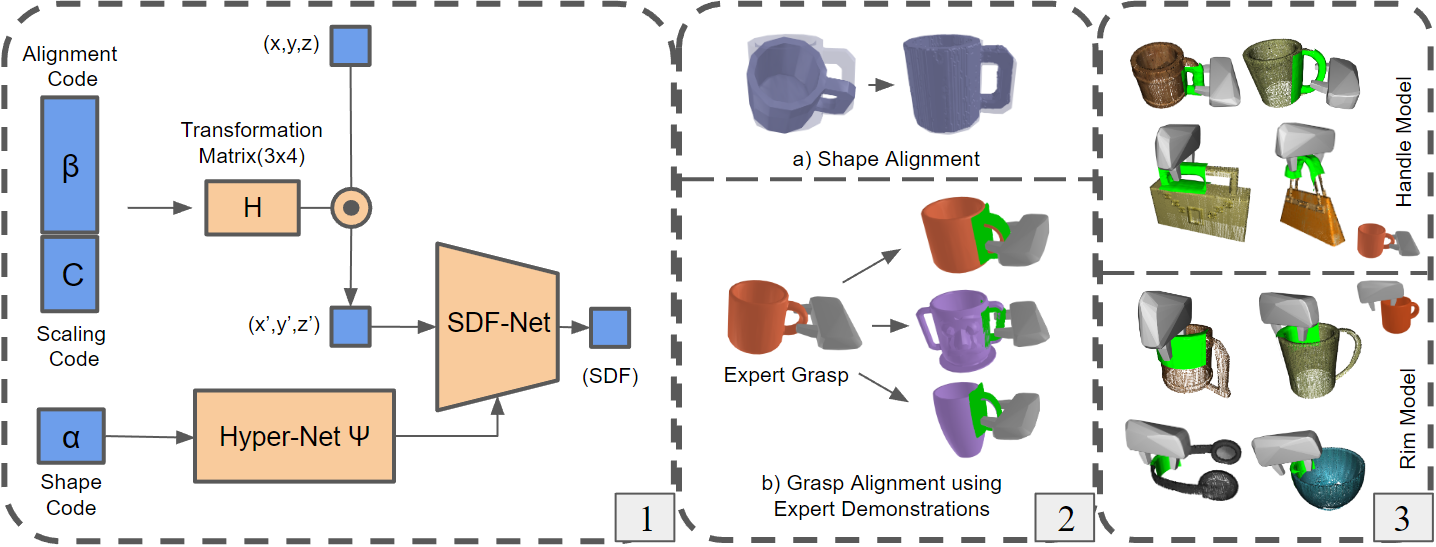}
    \caption{The proposed end-to-end network architecture, with self-aligning implicit representations (1). During Training (2), it simultaneously aligns objects with each other while learning to reconstruct their shape~(2-a). When this is done on local surfaces around the expert grasp demonstration, it aligns grasps within an object category (2-b). At inference time (3), learned models are used to identify and localize grasp locations on novel objects by finding surfaces similar to those on which the networks are trained on.}
    \label{fig:Architecture}
\end{figure*}

The overall architecture of the proposed approach is shown in Figure~\ref{fig:Architecture}. The proposed architecture learns a surface model while aligning the trained shapes to a shared common reference frame, as illustrated in step (2-a) in~Figure~\ref{fig:Architecture}. The shape alignment is localized to the surface around an expert grasp and is used to find the corresponding grasp locations on the rest of the training objects within the same category, as illustrated in step (2-b) in~Figure~\ref{fig:Architecture}. At inference time, the learned surface model is employed to transfer grasps to novel objects by identifying and localizing geometrically similar surfaces to the ones the model is trained on, as illustrated in step (3) in~Figure~\ref{fig:Architecture}. In this section, we will discuss the details of our approach.

\subsection{Preliminaries}
In our architecture, we build a signed-distance function (SDF)~\cite{deepsdf} based representation of local surfaces for shape reconstruction. SDF is an implicit function that represents the distance of its input point to the closest surface. SDF evaluates to negative values for the input points inside the object to be modeled, to positive values for the points outside the object and to $0$ for the points on the object surface.  

Our method consists of two MLP-based networks: SDF-Net, which predicts the SDF value for a queried position on the object, and Hyper-Net~\cite{hypernet}, which takes a shape code and predicts a part of the SDF-Net's weights. In our method, each reconstructed shape is assigned a shape code, and these shape codes are optimized together with the network weights during the training. In addition, we use position encodings, as they improve network performance on complex shapes~\cite{fourier_features}. Position encodings map 3D locations, $X$, to higher-dimensional inputs. We employ the positional encoding map $\Gamma(X) = [X,\Gamma_0(X),\Gamma_1(X),\ldots,\Gamma_{L-1}(X)]\in \mathbb{R}^{3+6L}$ where $\Gamma_{m}(X)=[\cos(2^m \pi X), \sin(2^m \pi X)] \in \mathbb{R}^{6}$. 

\subsection{Pose Alignment of Shapes}
The proposed architecture simultaneously finds 3D transformations that align the poses of local surfaces with each other while reconstructing their shapes. To train the system end-to-end, 3D transformations are modeled using $\mathfrak{se}(3)$ Lie algebra: $\mathbf{SE}(3)$ transformation has an associated Lie algebra $\mathfrak{se}(3)$. Using the exponential map $\mathfrak{se}(3)\rightarrow$ $\mathbf{SE}(3)$, we map 6-dimensional embeddings to transformations that continuously lie on $\mathbf{SE}(3)$ manifolds~\cite{SE3_tutorial}. Let $\beta = \langle\omega, t\rangle \in \mathfrak{se}(3)$  be a six-dimensional vector where $\omega,~t$ are three-dimensional vectors corresponding to rotation and translation. Equivalently, alignment transformation is presented by the matrix $T_{\text{alignment}} =  \left[\begin{array}{c|c}
R_{3 \times 3} &  v_{3 \times 1}\end{array}\right] \in \mathbf{SE}(3)$ with $v = V t$. 
Using the Rodrigues formula, $R$ and $V$ are written as
\begin{equation}
    R = e^{[\omega]} = I_3
+ \frac{\sin \theta}{\theta} [\omega]
+ \frac{1- \cos \theta}{\theta^2} [\omega]^2,
\end{equation}
\begin{equation}
    V = I_3
+ \frac{1 - \cos \theta}{\theta^2} [\omega]
+ \frac{1- \sin \theta}{\theta^3} [\omega]^2,
\end{equation}
where $[\omega]$ is the skew-symmetric matrix created from the  vector $\omega$, and $\theta = \|\omega\|$. To estimate the alignment transformation matrix $T_{\text{alignment}}$ from $\beta$, we use a Taylor-series expansion for linearization. This formulation is well defined, surjective, and allows for gradient-based optimization of $\beta$. At $\beta = 0$, $T_{\text{alignment}}$ corresponds to the identity transformation.

For scaling the objects we learn a scaling embedding $C\in\mathbb{R}^{3}$ that is then transformed to a $4\times 4$ transformation matrix $T_{\text{scaling}} = \mathrm{diag}([C, 1])$. Multiplying this matrix with $T_{\text{alignment}}$, we acquire an affine transformation matrix $H = T_{\text{alignment}}T_{\text{scaling}}$ that is used for pose and scale alignment. $H$ is used to transform a given $X$ location into $X'$ via $  X' = (H  \left[\begin{array}{c c}X & 1\end{array}\right]^T)^T $. $\beta$ and $C$ together are referred to as the pose refinement code. Optimizing the pose refinement code together with the shape code results in reconstructed surfaces that are well-aligned with each other. This is achieved without requiring any additional loss function that aids in the alignment of the poses of objects.

\paragraph*{Coarse-to-fine-Approximation}

Position embeddings improve the performance of coordinate-based neural networks by allowing them to represent higher frequency signals (\textit{i.e.,} more complicated shapes). However, their usage may lead to suboptimal performance on alignment as lower-frequency approximations of signals are better for alignment when there are larger pose differences. For this reason, by first approximating a shape using low-frequency bands of positional encoding and then progressively including higher-frequency bands, a coarse-to-fine approximation of the shape is performed. This process is achieved by applying a smoothing mask on the position encodings. More details can be found in~\cite{barf}. 

\subsection{Local Surface Modeling}

We employ local surface representation obtained from one object category to identify geometrically similar surfaces on novel objects. For training these local surface representations, we utilize object models from a single category, along with a reference frame provided on a representative instance within that category. This reference frame is positioned on the target local surface area. The remaining instances within this category will naturally possess local surfaces that are geometrically similar to the targeted one around this reference frame. These surfaces are then utilized to train the implicit surface model.

We sample points exclusively within a sphere of radius $r$ centered on the reference frame, which restricts the alignment to the designated area within the sphere, excluding any other surface areas. It also has the additional benefit of limiting the network's capacity spent on surfaces outside of the targeted one. Since the alignment transformation is optimized at the same time with the network parameters, the sampling sphere is dynamically adapted to the current transformation matrix $H$. This ensures that only the targeted surface is learned during training.

\subsection[Grasp Transfer]
{Grasp Transfer~\footnote{We use the transformation notation ${}^{o}T_{SD}$, where S is the source frame, $D$ is the destination frame, and $o$ is the corresponding object.}
}

We associate demonstrated grasps with local surface models, and use these models to identify geometrically similar surfaces on novel objects for grasp transfer. To achieve this, the reference frames for training these models are provided through grasp demonstrations. 

We first record an expert grasp demonstration on an object, which we assign as the anchor object (a). Using this expert grasp demonstration, we find a transformation between the world frame (W) and the demonstration frame (D) which will be equal to the transformation between the world frame and the grasp frame (G) for the anchor object $T_{WD} ={}^{a}T_{WG}$. Using $T_{WD}$, we transform all object frames to the demonstration frame, which will put their grasp frames close to the vicinity of their current identity pose. 

Our method then learns to reconstruct each local surface while simultaneously finding the transformations that align the shapes ($i$ denotes the index of the shape) from the demonstration frame to a common reference frame denoted as the alignment frame (A) $ {}^{i}H = {}^{i}T_{DA}$. When local surfaces are well-aligned, we assume that the transformation between the alignment frame and the grasp frame will be independent of the object, and can be denoted as $T_{AG}$. The transformation between the alignment and the grasp frame is found using the alignment transformation of the anchor object $T_{WD} {}^{a}T_{DA} T_{AG}={}^{a}T_{WG}=T_{WD}\implies T_{AG}={}^{a}T_{AD}$. For the remaining objects, we estimate the transformation between the world frame and the grasp frame as ${}^{i}T_{WG} =  T_{WD} {}^{i}T_{DA} T_{AG}$. 

After the learning step, for transferring the grasp to new objects, we use the learned implicit representations. During the inference step, the weights of SDF-net and Hypernet are fixed. We pick several candidate frames, with transformations ${}^{i}T_{WC_j}$ ($C_j$ corresponds to the candidate frame with index $j$), at locations close to the target surface for the object $i$ with a local surface that is geometrically similar to the originally trained one. We employ these candidate frames on copies of the object to put them in the vicinity of possible grasp locations. We then optimize a shape code along with a pose refinement code for each object copy to identify surfaces similar to the underlying surface type. After the optimization, if the reconstruction error is lower than a given threshold, the pose refinement code corresponding to the candidate $j$ of the object $i$ is used to estimate the transformation matrix ${}^{i,j}H$ which denotes the transformation between the candidate frame and the alignment frame ${}^{i,j}H={}^{i}T_{C_jA}.$ The transformation between the candidate frame and the alignment frame is employed together with the transformation between the world frame and the candidate frame to identify grasp poses on new objects ${}^{i}T_{WG}  =  {}^{i}T_{WC_j} {}^{i}T_{C_jA} T_{AG}$.

\subsection{Local Surface Prediction on Point clouds}

At inference time, we work with point cloud observations and build their corresponding SDF representation as the first step in the local surface prediction stage. Point clouds have SDF values of $0$, as the observed points are on the surface of the objects. While these points can be used directly for optimizing shape and pose refinement codes, it often gives sub-optimal results. Instead, we sample points uniformly in the workspace and find their unsigned distance value directly by checking the distance between them and the originally observed point cloud. Using these sampled points in addition to the observed point cloud in optimization yields better shape reconstruction and grasp pose estimation.

\section{Experiments}

We evaluate our method based on three simulation experiments: shape alignment, grasp alignment and grasp transfer. Furthermore, we validate our method using a real robot, showing that it can transfer grasps to novel real objects, such as the watering pot and the bowl shown in Figure~\ref{fig:real_world_alignment}, where we use the same models trained with synthetic objects, \textit{e.g.,} mugs.

We use the ShapeNet-Core V2 dataset~\cite{shapenet} which has all the shapes in their canonical poses. We employ the same data generation process as DeepSDF~\cite{deepsdf}; however, as we deal with smaller surface areas, we perturb surface points with a higher variance accordingly ($0.0025 \times 15$). In all experiments, we use a 5-layer ReLU MLP for SDF-Net and multiple 2-layer ReLU MLPs for Hypernet. All MLPs have a hidden layer size of 256. We use Adam optimizer~\cite{adam} with a learning rate of $10^{-3}$. The loss function we used for training is $L_T = w_1 L_{SDF} + w_2 L_{shape} + w_3 L_{tr}$ in which $L_{SDF}$ is the $L1$-distance between the predicted and the ground truth SDF values; $L_{shape}$ and $L_{tr}$ are the $L2$ norm of the shape code, $\alpha$, and translation vector, $t$, respectively. At inference time, an additional loss term for scaling code, $L2$ norm of $C-1$, with weight $w_4$ is added to constrain it~\footnote{In this work, $w_1=3 \times 10^{3}$, $w_2= 10^{4}$, $w_3=10^{2}$, $w_4=2.5 \times 10^2$}.  
Experiments are performed using a system equipped with an Intel i9-11950H processor (2.6 GHz, 8 cores), an Nvidia RTX A3000 GPU and 32 GB of RAM. The training of a single shape model typically takes approximately 25 minutes. 

For grasp pose estimation, we train three implicit surface models that represent local surface types that are frequently encountered during two-finger grasping. Each model is trained with local surfaces extracted from 64 object meshes. We employ a single grasp pose demonstration as a reference for the rim and the handle of the anchor mug (orange mug shown in Figure~\ref{fig:Architecture}) to train the rim and the handle grasp models. We train a cylinder grasp model based on primitive cylinders and cuboids with varying sizes~\footnote{Cylinders with diameters $\varnothing$ = [$10$, $80$] cm and heights h = [$25$, $100$] cm, cuboids with side lengths x = [$10$, $80$] cm and heights h = [$25$, $100$] cm.}. For these objects, the reference grasp pose is given on the sides of cuboids and cylinders, and grasp poses are pre-aligned. To allow the model to represent cuboids and cylinders of different sizes, the pose refinement codes are fixed during the training of this model. 

We perform the simulation experiments using a Panda robot. In real-world experiments, our experimental setup consists of a UR10 robotic arm equipped with an RG-2FT gripper. In both cases, we use a Real-Sense D435 camera to acquire the point cloud of the scene. The camera is fixed on the wrist of the robotic arm. In the experiments, all objects are placed in stable standing poses on a table top.

\subsection{Shape Alignment}

\begin{figure}[t]
    \centering
    \includegraphics[width=0.90\linewidth]{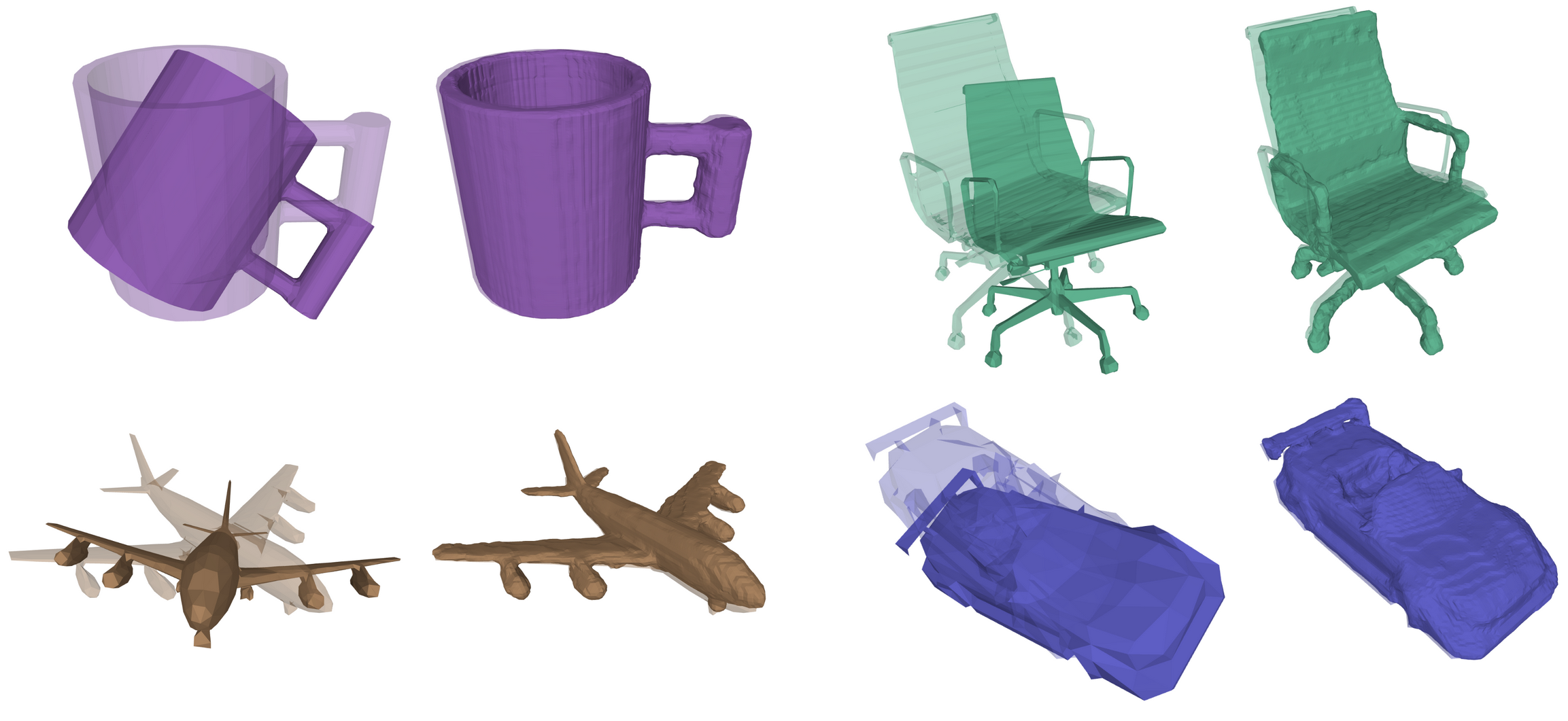}
    \caption{Example shape alignment results. The meshes on their canonical poses are shaded. Reconstructed meshes are on the right, given perturbed meshes are on the left. Post-training, shapes are aligned very closely to their canonical poses. }
    \label{fig:alignment}
\end{figure}

In this experiment, we show that our network can align the poses of objects from a specific category throughout the training. We use 64 shapes from four different object categories, which are mugs, chairs, planes, and cars, to create four datasets. For each shape in the datasets, the poses of the objects are varied by applying random perturbations. We train our network with one object in its canonical pose as the anchor, and the rest of the objects with randomly perturbed positions and orientations. In this experiment, scaling embeddings are fixed as they cannot be used in the evaluation.

Visualizations of example reconstructions are shown in Figure~\ref{fig:alignment}. Our method aligns objects very close to their canonical poses. We have evaluated our method's alignment performance by comparing two different Chamfer distances for all the object categories. The first Chamfer distance is between the perturbed mesh and the ground-truth mesh, and the second is between the reconstructed mesh and the ground-truth mesh. For this, we sample 10,000 points from all meshes (except for the anchor one) and estimate the two-way Chamfer distance. We repeat this experiment five times. The results are shown in Table~\ref{table_alignment}. This table shows that the Chamfer distances for reconstructed meshes are significantly lower than the Chamfer distances for perturbed poses. Another observation is that our method exhibits a higher reconstruction loss and standard deviation for chairs and planes. This disparity is due to significant variations in terms of geometry within these categories, e.g. the chair category not only includes chairs but also sofas. Regardless, even for these categories, our method acquires lower Chamfer distances. Overall, these results show that our approach can align object meshes while learning their implicit representations.

\begin{table}[t]
\caption{Shape Alignment Results} 
\label{table_alignment}
\centering
\begin{tabular}{c||c|c}
\bfseries Object Name  & \bfseries Reconstructed Mesh & \bfseries Pose Perturbed Mesh  \\
\hline\hline
Mugs & $1.5 \pm 0.5 $ & $17.5 \pm 1.0$\\ 
Chairs & $5.8 \pm 3.9 $ & $20.7 \pm 1.0$ \\
Planes & $7.8 \pm 3.7 $ & $25.2 \pm 3.9$ \\
Cars & $0.8 \pm 0.3 $ & $20.1 \pm 2.4$ \\
\end{tabular}
\begin{tablenotes} 
\item $^*$ The error values in the table are divided by $10^{-3}$. $\pm$ denotes the standard deviation.
\end{tablenotes} 
\end{table}

\subsection{Spatial Precision of Grasp Alignments}

To evaluate the grasp alignment performance of our method, we compare it to a recent baseline, neural descriptor fields~(NDF) \cite{NDF}, based on two metrics. The first metric is spatial precision of pre-grasp positions, \textit{i.e.,} the distance between grasps generated by our method and the demonstrated grasp. This metric is calculated based on the Euclidean distances between the predicted and the expert annotated grasp poses. The second metric is grasp success accuracy in simulation via lifting tests by checking whether the object slips or it is dropped.  This experiment is conducted on the handle region of the mug objects as this part is unique for mugs~\footnote{We also tested NDF with the multi-category model. However, the model did not exhibit consistent success in predicting grasp locations correctly. Hence, we limited our baseline comparison to mug objects. Furthermore, we perform this comparison solely on points sampled from the CAD model of an object as the NDF did not work very well with the point cloud acquired from the simulation. This is most likely due to the fact that such point cloud is partial, \textit{i.e.,} some parts of the mug are not visible to the camera.}.

\begin{table}[t]
\caption{Spatial Precision and Grasp Accuracy ($\%$)} 
\renewcommand{\arraystretch}{0.9}
\setlength{\tabcolsep}{5pt}
\begin{tabular}{c||c|c|c|c|c|c}
\textbf{$\epsilon$}  & \multicolumn{2}{|c|}{\textbf{Ours}} &  \multicolumn{2}{|c|}{\textbf{NDF~\cite{NDF}}} & \multicolumn{2}{|c}{\textbf{Naive-Transfer}} \\\hline\hline
$0^\circ$&$\mathbf{0.16 \pm 0.1}$&$ \mathbf{100}$ &$0.29 \pm 0.2  $&$ \mathbf{100}$ &$0.18 \pm 0.1  $&$ \mathbf{100}$ \\
$10^\circ$&$\mathbf{0.15 \pm 0.1}$&$ \mathbf{100}$ &$0.42 \pm 0.2 $&$ 88$ &$0.30 \pm 0.1  $&$ \mathbf{100}$ \\
$20^\circ$&$\mathbf{0.16 \pm 0.1}$&$ \mathbf{100}$ &$0.30 \pm 0.1  $&$ \mathbf{100}$ & $0.52 \pm 0.1  $&$ \mathbf{100}$ \\
$30^\circ$&$\mathbf{0.16 \pm 0.1}$&$ \mathbf{100}$ &$0.53 \pm 0.8 $&$ 88$ &$0.76 \pm 0.1 $&$ 0$ \\
$40^\circ$&$0.47 \pm 0.7 $&$ 81$ &$\mathbf{0.43 \pm 0.2}  $&$ \mathbf{94}$ &$1.00 \pm 0.1 $&$ 0$ \\
\end{tabular}
\label{baseline_comparison}
\begin{tablenotes} 
\item $^*$ $\pm$ denotes standard deviation.
\end{tablenotes} 
\end{table}

\begin{figure}[t]
    \centering\small
    \includegraphics[width=0.95\linewidth]{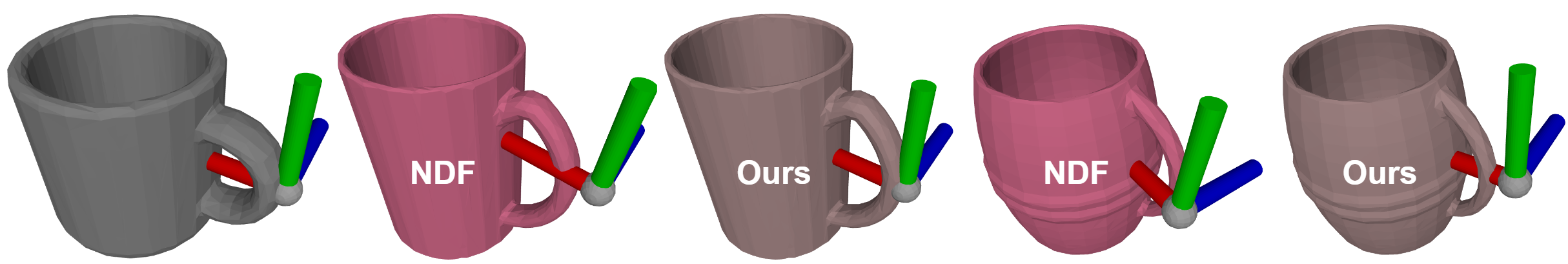}
    \caption{Comparison of resulting grasp poses. The anchor mug with expert grasp pose is shown with the gray mesh on the left. Grasp poses, represented as fingertip frames, predicted by NDF and our method are shown on the pink and the brown meshes.
    }
    \label{fig:spatial_precision}
\end{figure}

We annotate expert grasp locations on 16 mug objects to compare the proposed network with NDF. NDF transfers the grasp to objects of the same category with energy minimization on the latent space of queried positions. We utilize the pose of the same expert grasp demonstration on the anchor object used in training our method as a query to condition the NDF model. For both models, initial grasp candidate locations are generated around the expert grasp pose. We repeat this experiment with four different levels of perturbations applied to object yaw angle: $10^\circ,20^\circ,30^\circ,40^\circ$ in which we rotate the test objects by the corresponding degree before optimizing the grasp locations. As an additional baseline, we include the naive transfer results where the originally demonstrated grasp on the anchor object is directly transferred without any adjustments. Results of these experiments can be found in Table~\ref{baseline_comparison}. Results are normalized to have a maximum error of 1. Illustrations of two example grasps predicted by our method and NDF can be found in Figure~\ref{fig:spatial_precision}. Table~\ref{baseline_comparison} and Figure~\ref{fig:spatial_precision} show that our method predicts grasps more similar to the expert ones until $40^\circ$, and acquires better spatial precision than NDF. At $40^\circ$, it still predicts good grasps for most of the objects but fails for some of them potentially due to getting stuck in a local minima. Compared to our method, NDF has a smaller error at $40^\circ$ as it does not need as good initialization as our method does for grasp candidate sampling, however, it overall performs worse than our method.  Furthermore, our method leads to an average optimization time of $4.62$ seconds for 8 grasp pose candidates while NDF requires $62.2$ seconds for the same number of poses. Grasp pose timings for our network grow linearly with the number of grasp candidates.

\subsection{Grasp Transfer}
In this experiment, we show that the learned implicit representations can be used in identifying grasps on unseen mugs and novel objects from different categories: bags, bowls, earphones, bottles, and cans. For this, each object is placed in a stable pose on a table in PyBullet \cite{pybullet} simulation.  A camera mounted on the wrist captures point cloud observations of the object from multiple predetermined viewpoints. These point clouds are then merged and filtered to retain only the points of the object of interest.

\begin{table*}[t]
\centering
\caption{\textbf{Grasp Success Rate ($\%$)~/~Post Grasp Motion (cm)}}
\begin{tabular}{l||c|c|c|c|c|c|c|c|c|c|c|c}

\textbf{Object~-~Part Name~(\# of objects)}  & \multicolumn{2}{c}{\textbf{Ours}} &  \multicolumn{2}{|c|}{\textbf{NOCS-$0^\circ$}} & \multicolumn{2}{|c|}{\textbf{NOCS-$20^\circ$}} &  \multicolumn{2}{|c|}{\textbf{NOCS-$40^\circ$}} &  \multicolumn{2}{|c|}{\textbf{NOCS-$60^\circ$}} & \multicolumn{2}{|c}{\textbf{NOCS-$80^\circ$}} \\\hline\hline
Mugs - Handle~(16) & $\mathbf{100}$&$\mathbf{0.2}$ & $\mathbf{100}$ & $0.58$ & $100$&$0.7$ & $12$&$1.66$ & $0$&$NA$ & $0$&$NA$ \\
Mugs - Rim~(16) & $\mathbf{100}$&$\mathbf{0.41}$ & $\mathbf{100}$&$0.57$ & $100$&$0.66$ & $100$&$1.32$ & $100$&$1.94$ & $100$&$2.3$ \\
Bags - Handle~(23) & $87$&$\mathbf{0.2}$ & $\mathbf{100}$&$0.73$ & $91$&$1.02$ & $83$&$1.57$ & $30$&$2.67$ & $4$&$0.25$ \\
Bowls - Rim~(65) & $\mathbf{98}$&$\mathbf{0.66}$ & $\mathbf{100}$&$1.06$ & $98$&$1.09$ & $98$&$1.14$ & $100$&$1.14$ & $97$&$1.07$ \\
Earphones - Rim~(22) & $91$&$0.69$ & $\mathbf{100}$&$\mathbf{0.62}$ & $100$&$1.63$ & $95$&$2.84$ & $68$&$3.53$ & $45$&$3.48$ \\
Bottles - Cylinder~(35) & $\mathbf{94}$&$\mathbf{0.59}$ & $80$&$0.67$ & $77$&$0.76$ & $71$&$0.65$ & $80$&$0.99$ & $80$&$0.62$ \\
\end{tabular}
\label{grasp_accuracy_success}
\end{table*}

\begin{figure}[t]
    \centering
    \includegraphics[width=0.95\linewidth]{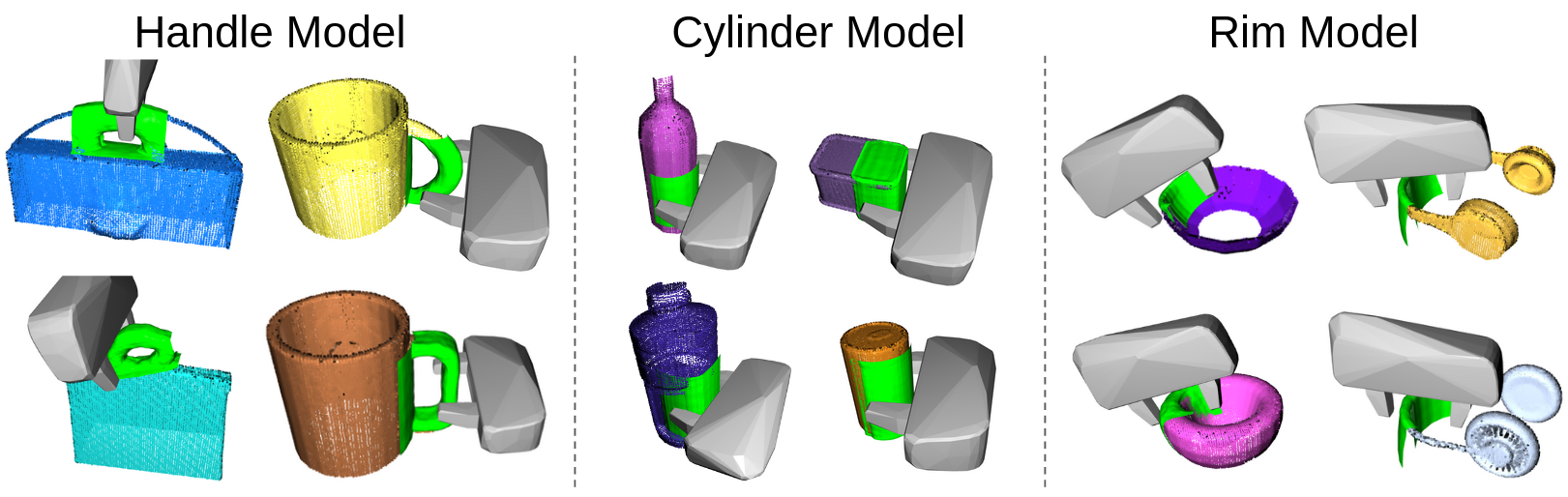}
    \caption{Results of transferring grasps to different objects. Reconstructed surfaces are shown with green meshes. 
    }
    \label{fig:grasp_transfer_sim}
\end{figure}

\begin{figure*}[t]
	\centering
    \includegraphics[width=0.72\linewidth]{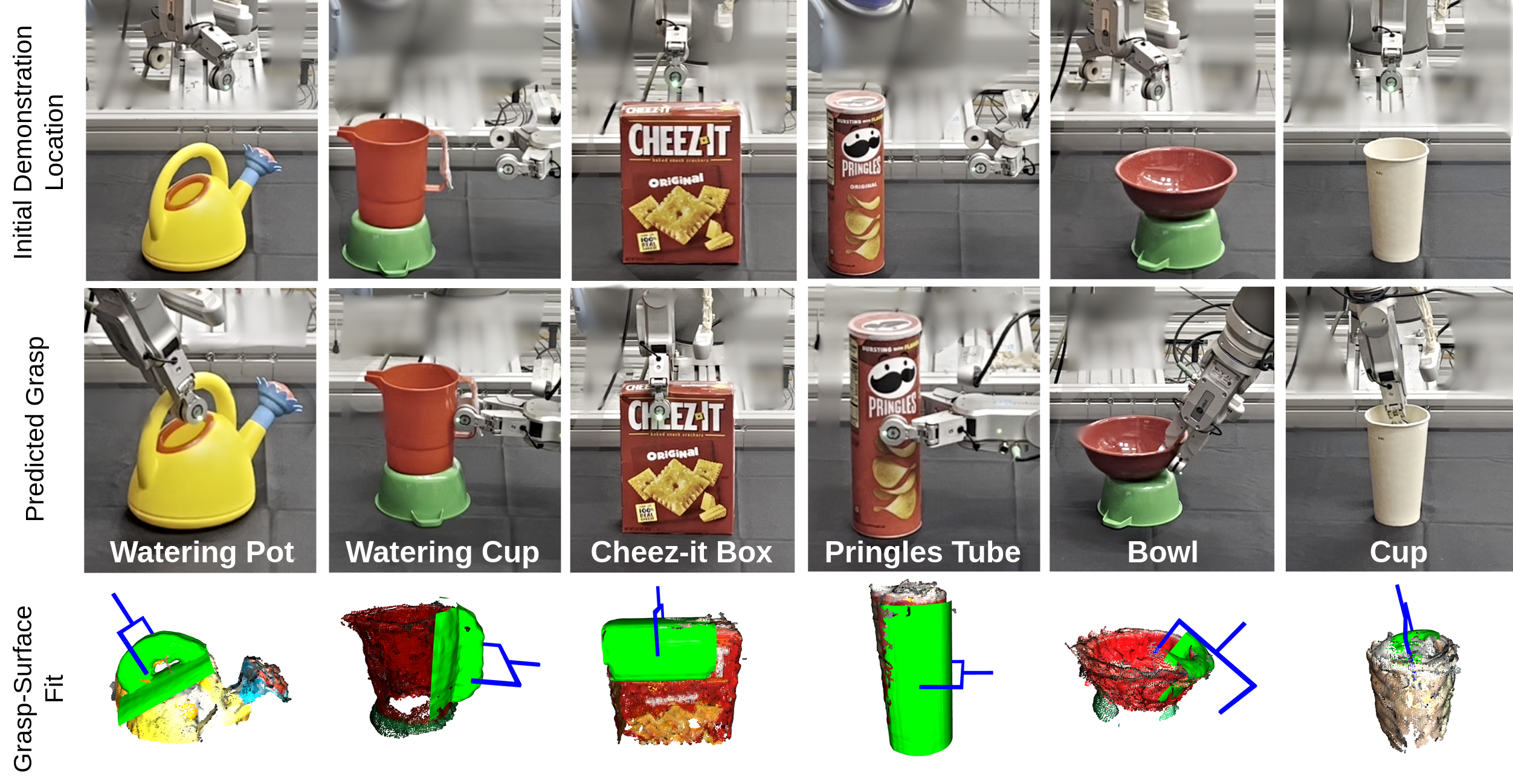}
    \caption{Grasp pose alignments on real-world point clouds. Our method leverages a single rough initial demonstration for each object and finds the correct grasp pose by searching for a grasp-surface fit.}
    \label{fig:real_world_alignment}
\end{figure*} 

For baseline comparison, we use the normalized object coordinate system (NOCS)~\cite{nocs} for grasp pose transfer. In NOCS, all points in the point cloud are normalized uniformly such that each of their $x$, $y$, $z$ values are between $0$ and $1$ with minimum and maximum being $0$ and $1$. For the baselines, depending on the object type, the grasp pose from the handle, rim, and cylinder categories are defined in NOCS coordinates. For this, expert grasp poses on a single representative instance from each object type are used. NOCS-based grasp pose estimation requires accurate object pose detection. Therefore, we devise five baselines where we assume NOCS methods have access to canonical object poses with yaw angle errors of $0^{\circ}$, $20^{\circ}$, $40^{\circ}$, $60^{\circ}$, and $80^{\circ}$. Our method also has access to the provided grasp pose for each object type for grasp candidate sampling, along with the corresponding surface type, however, it does not have access to the object poses. Specifically, utilizing only the provided object-level grasp pose locations, we sample 12 candidate grasp poses with $30^\circ$ intervals around the object and optimize their parameters.

We run our method and the NOCS baseline on the point clouds of each object for grasp transfer. Visualizations of some of the grasp poses can be seen in Figure~\ref{fig:grasp_transfer_sim}. The grasp transfer approach is assessed using two criteria. The first criterion is based on post grasp motion, \textit{i.e.,} the Euclidean distance between the object's position before and after the robot executes its grasping action, prior to lifting the object. The second criterion is based on whether or not the object can be successfully lifted up after the grasp execution.

The grasp accuracy and the post-grasp motion can be found in Table~\ref{grasp_accuracy_success}. In Table~\ref{grasp_accuracy_success}, we observe that our method leads to successful grasp transfer results on all experiment objects with similar success rates to NOCS-$0^\circ$. When the amount of angle error for the NOCS baseline increases, our network outperforms the NOCS baseline, especially on non-symmetric objects. One interesting observation here is that our method always performs better than the NOCS baseline on bottle objects even though they are mostly symmetric. After analyzing grasps on bottle category, we observe that bottle objects in their canonical pose do not consistently have the geometrically shorter side in front. This leads to failure for NOCS-based baselines as they rely on consistency in geometry for objects in their canonical pose. As a result, the NOCS-based methods sometimes grasp bottles from their longer side. Our method however adapts to shape variations and consistently predicts grasps from the shorter side of the bottle. In addition, as our method generates grasp poses based on local surface knowledge, it acquires lower post-grasp motion than the NOCS baselines, even on objects in which pose perturbations do not hinder the grasp success.

\subsection{Real Robot Experiments}

In these experiments, we analyze our method's performance using a real robot composed of a UR10 arm and an OnRobot RG2-FT gripper, and six objects as seen in Figure~\ref{fig:real_world_alignment}. We demonstrate that our approach can successfully transfer grasp poses using point cloud observations from real objects belonging to unseen categories. This is achieved by utilizing the same models trained with synthetic data. For each object, we initialize our method by providing a single demonstration, which is around the surface area to be grasped for grasp candidate sampling, and the surface type. The robot scans the scene from six predefined camera views and merges the acquired point clouds. The camera views are chosen independently of the object's poses. Then the initial demonstration is utilized in grasp candidate sampling to find the correct surface-grasp alignment on the merged point cloud. Using this alignment, a grasp is executed. Following these steps, we perform grasp transfer experiments for each object five times using varying object poses to test robustness to pose changes. We use the same initial demonstration for all trials involving each corresponding object. For each grasp execution, scene capture takes $1$ minute and grasp pose estimation takes, on average, around $10.5$ seconds.

Our method's grasp predictions on the objects are shown in Figure~\ref{fig:real_world_alignment}. The success rates for watering pot, watering cup, Cheez-it box, Pringles tube, bowl, and cup are as follows: {5/5, 4/5, 5/5, 5/5, 5/5, 5/5}, respectively. Overall, the method demonstrates high success rates in transferring grasps to real novel objects including cases where the objects are partially visible, with only one observed failure. In the failure case, the final transferred grasp pose leads to successful lifting of the object, but it is counted as failure. The actual reason for failure is that the best handle surface fit is found at the rim of the object, not at the handle, which might be due to the combination of the stochasticity in the initial grasp candidate sampling and the point cloud being noisy.

\section{Conclusion}

This paper presents an implicit representation architecture that aligns local surfaces around an expert grasp demonstration while learning to reconstruct their shapes. This representation is used to transfer grasps to local surfaces that are similar in shape to the surface the expert grasp is demonstrated on. Through both simulation and real-world experiments, we show that we can transfer grasps to new objects with similar local geometry and acquire better spatial precision than a baseline approach. 

In future work, we plan to investigate the effects of employing equivariant shape representations \cite{se3_equiv, condor}, keypoint detection methods, and zero-shot category level pose estimators~\cite{goodwin2022zero} in terms of improving the generation and evaluation of local surface candidates. In addition, we aim to perform faster grasp pose generation and scale our method to multi-object scenarios~\cite{murali20206}. We also plan to use multi-fingered hands for a direct mapping between fingertip locations and the local surfaces.





\bibliographystyle{IEEEtran}
\bibliography{IEEEabrv,references}

\end{document}